\documentclass[utf8]{frontiersSCNS} 

\usepackage{url,hyperref,lineno,microtype,subcaption}
\usepackage[onehalfspacing]{setspace}

\linenumbers

\def\keyFont{\fontsize{8}{11}\helveticabold }
\def\firstAuthorLast{Gershenson} 
\def\Authors{Carlos Gershenson\,$^{1,2,3,*}$}
\def\Address{$^{1}$Instituto de Investigaciones en Matemáticas Aplicadas y Sistemas,
Universidad Nacional Autónoma de México, Mexico City, 04510, Mexico \\
$^{2}$Centro de Ciencias de la Complejidad,
Universidad Nacional Autónoma de México, Mexico City, 04510, Mexico \\ 
$^{3}$ Lakeside Labs GmbH, Lakeside Park B04, 9020 Klagenfurt am Wörthersee, Austria}
\def\corrAuthor{ Carlos Gershenson\\
	Instituto de Investigaciones en Matemáticas Aplicadas y en Sistemas\\
	Universidad Nacional Autónoma de México\\
	A.P. 20-126, 01000, México, CDMX, México}

\def\corrEmail{cgg@unam.mx}

\begin{document}
\onecolumn
\firstpage{1}

\title[Intelligence as information processing]{Intelligence as information processing: brains, swarms, and computers} 

\author[\firstAuthorLast ]{\Authors} 
\address{} 
\correspondance{} 

\extraAuth{}

\maketitle

\begin{abstract}

\section{}
There is no agreed definition of intelligence, so it is problematic to simply ask whether brains, swarms, computers, or other systems are intelligent or not. To compare the potential intelligence exhibited by different cognitive systems, I use the common approach used by artificial intelligence and artificial life: Instead of studying the substrate of systems, let us focus on their organization. This organization can be measured with information. Thus, I apply an informationist epistemology to describe cognitive systems, including brains and computers. This allows me to frame the usefulness and limitations of the brain-computer analogy in different contexts. I also use this perspective to discuss the evolution and ecology of intelligence.

\tiny
 \keyFont{ \section{Keywords:} mind, cognition, intelligence, information, brain, computer, swarm} 
\end{abstract}

\section{Introduction}

In the 1850s, an English newspaper described the growing global telegraph network as a ``nervous system of the planet'' \citep{gleick2011information}. Notice that this was half a century before Santiago Ram\'on y Cajal~\citeyearpar{ramon1899textura} first published his studies on neurons. Still, metaphors have been used since antiquity to describe and try to understand our bodies and our minds~\citep{Zarkadakis2015,Epstein2016}: humans have been described as made of clay (Middle East) or corn (Americas), with flowing humors, like clockwork automata, similar to industrial factories, etc. The most common metaphor in cognitive sciences has been that of describing brains as computers \citep{von-Neumann1958,10.3389/fcomp.2021.681416}.

Metaphors have been used in a broad range of disciplines. For example, in urbanism, there are arguments in favor of changing the dominant narrative of ``cities as machines'' to ``cities as organisms'' \citep{Batty:2012Cities,Gershenson:2013}.

We can have a plethora of discussions on which metaphors are the best. Still, being pragmatic, we can judge metaphors in terms of their usefulness: if they help us understand phenomena or build systems, then they are valuable. Notice that then, depending on the context, different metaphors can be useful for different purposes \citep{Gershenson2004}. For example, in the 1980s, the debate between symbolists/representationists (brain as processing symbols) \citep{fodor1988connectionism} and connectionists (brain as network of simple units) \citep{smolensky1988proper} did not end with a ``winner'' and a ``loser'', as both metaphors (computational, by the way) are useful in different contexts.

There have been several other metaphors used to describe cognition, minds, and brains, each with their advantages and disadvantages \citep{Varela1991,steels1995artificial,clark1998extended,Beer2000,Gardenfors2000,Garnier:2007,Froese2009,Kiverstein2009,Chemero2009,Enaction2010,Froese2010,Downing2015Intelligence-Em,doi:10.1177/1059712319856882}. It is not my purpose to discuss these here, but to notice that there is a rich variety of flavors when it comes to studying cognition. Nevertheless, all of these metaphors can be described in terms of information processing. Since computation can be understood as the transformation of information \citep{Gershenson:2007}, ``computers'', broadly understood as machines that process information can be a useful metaphor to contain and compare other metaphors. Note that the concept of ``machine'' (and thus computer) could also be updated \citep{10.3389/fevo.2021.650726}.

Formally, computation was defined by Turing~\citeyearpar{Turing:1937}. A computable function is that which can be calculated by a Universal Turing Machine (UTM). Still, there are two main limitations of UTMs related to modeling minds \citep{Gershenson2011}:
\begin{enumerate}
\item \textbf{UTMs are closed}. Once a computation begins, there is no change in the program or data, so adaptation during computation is limited. 
\item \textbf{UTMs compute only once they halt}. In other words, outputs depend on a UTM ``finishing its computation''. Still, minds seem to be more continuous than halting. Then the question arises: what function would a mind be computing?
\end{enumerate}

As many have noted, the continuous nature of cognition seems to be closely related to that of the living \citep{Maturana1980,Hopfield1994,Stewart1995,Walker2014}. We have previously studied the ``living as information processing'' \citep{Farnsworth2013Living-is-Infor}, not only at the organism level, but at all relevant scales. Thus, it is natural to use a similar approach to describe intelligence.

In the next section, I present a general notion of information and its limits to study intelligence. Then, I present the advantages of studying intelligence in terms of information processing. Intelligence is not restricted to brains, and swarms are a classic example of this, which can also be described as information processing systems. Before concluding, I exploit the metaphor of ``intelligence as information processing'' to understand its evolution and ecology.

\section{Information}

Shannon~\citeyearpar{Shannon1948} proposed a measure of information in the context of telecommunications, that is equivalent to Boltzmann-Gibbs entropy. This measure characterizes how much a receiver ``learns'' from incoming symbols (usually bits) of a string, based on the probability distribution of previously known/received symbols: if new bits can be completely determined from the past (as in a string with only one repeating symbol), then they carry zero information (because we know that the new symbols will be the same as previous ones). If previous information is useless to predict the next bit (as in a random coin toss), then the bit will carry maximum information. Elaborating on this, Shannon calculated how much redundancy is required to reliably transmit a message over an unreliable (noisy) channel. Even when Shannon's purpose was very specific, the use of information in various disciplines has exploded in recent decades \citep{Haken1988,Wheeler1990,Lehn:1990,GellMann:1996,AtlanCohen1998,DeCanioWatkins1998,vonBaeyer2004,Roederer2005,CoverThomas2006,ProkopenkoEtAl2007,Prokopenko:2011,Batty:2012,Escalona-Moran:2012,Gershenson:2007,Fernandez2013Information-Mea,Zubillaga2014Measuring-the-C,Hidalgo2015,Haken2015Information-Ada,Murcio2015,Amoretti2015Measuring-the-c,FernandezCxLakes,Roli2018,Krakauer:2020,10.7717/peerj.8533,Gershenson2018,10.1162/isal_a_00402,Scharf2021}.

We can say that electronic computers process information \emph{explicitly}, as we can analyze each change of state and information is encoded in a precise physical location. However, humans and other animals process information \emph{implicitly}. For example, we say we have memories, but these are not physically at a specific location. And it seems unfeasible to represent precisely the how information changes in our brains. Still, we do process information, as we can describe ``inputs'' (perceptions) and ``outputs'' (actions).

Shannon assumed that the \emph{meaning} of a message was agreed previously between emitter and receiver. This was no major problem for telecommunications. However, in other contexts, meaning is not a trivial matter. Following Wittgenstein~\citeyearpar{Wittgenstein1999}, we can say that the meaning of information is given by the \emph{use} agents make of it. This has several implications. One is that we can change meaning without changing information (passive information transformation \citep{Gershenson:2007}). Another is the limits on artificial intelligence \citep{searle1980minds,mitchell2019artificial}, as the \emph{use} of information in artificial systems tends to be predefined. Algorithms can ``recognize'' traffic lights or cats in an image, as they are trained for this specific purpose. But the ``meaning'' for computer programs is predefined, \emph{i.e.} what we want the program to do. The quest for an ``artificial general intelligence'' that would go beyond this limit has produced not much more than speculations.

Even if we could simulate in a digital computer all the neurons, molecules, or even elementary particles from a brain, such a simulation would not yield something akin to a mind. On the one hand, \emph{interactions} generate novel information at multiple scales. On the other hand, as mentioned above, \emph{observers} can give different meanings to the same information. In other words, the same ``brain state'' could refer to different ``mental state'' for different people. In a sense, this is related to the failure of Laplace's daemon: even with full information of the states of the components of a system, prediction is limited because interactions generate novel information \citep{Gershenson:2011e}. And this novel information can determine the future production of information at different scales through upward or downward causation \citep{Campbell1974,Bitbol2012,Flack2017}, so all relevant scales should be considered \citep{Gershenson2021:emergence}.

In spite of all its limitations, the computer metaphor can be useful in a particular way. First, the limits on prediction by interactions are related to \emph{computational irreducibility} \citep{Wolfram:2002}. Second, describing brains and minds in terms of information allows us to avoid dualisms. Thus, it becomes natural to use information processing to describe intelligence and its evolution. Finally, information can contain other metaphors and formalisms, so it can be used to compare them and also to exploit their benefits.

\section{Intelligence}

There are several definitions of intelligence, but not a single one that is agreed upon. We have similar situations with the definitions of life \citep{DeDuve2003,Aguilar2014The-Past-Presen}, consciousness \citep{Michel2019},  complexity \citep{lloyd2001measures,HeylighenEtAl2007}, emergence \citep{bedau2008emergence},
and more. These concepts could be said to be of the type ``I know it when I see it'', to quote Potter Stewart.

Still, having no agreed definition is no motive nor excuse for not studying a phenomenon. Moreover, having different definitions for the same phenomenon can give us broader insights than if we stick to a single, narrow, inflexible definition.

Thus, we could define intelligence as ``the art of getting away with it'' (Arturo Frapp\'e), or ``the ability to hold two opposed ideas in mind at the same time and still retain the ability to function. One should, for example, be able to see that things are hopeless and yet be determined to make them otherwise” (F. Scott Fitzgerald). Alan Turing~\citeyearpar{Turing:1950} proposed his famous test to decide whether a machine was intelligent. Generalizing Turing's test, Mario Lagunez suggested that in order to decide whether a system was intelligent, first, the system has to perform an action. Then, an observer has to \emph{judge} whether the action was intelligent or not, according to some criteria. In this sense, there is no intrinsically intelligent behavior. All actions and decisions are contextual \citep{Gershenson2002ua}. Like with meaning, the same action can be intelligent or not, depending on the context and on the judge and their expectations. 

Generalizing, we can define intelligence in terms of information processing: An agent $a$ can be described as intelligent if it transforms information (individual (internal) or environmental (external)) to increase its ``satisfaction'' $\sigma$.

I have previously defined satisfaction $\sigma \in [0,1]$ as the degree to which the goals of an agent have been fulfilled \citep{GershensonDCSOS,Gershenson:2010a}. Certainly, we still require an observer, since we are the ones who define the goals of an agent, its boundaries, its scale, and thus, its satisfaction. Examples of goals are sustainability, survival, happiness, power, control, and understanding. All of these can be described as \emph{information propagation} \citep{Gershenson:2007}: In this context, an intelligent agent will propagate its own information. 

Brains by themselves cannot propagate. But species of animals with brains tend to propagate. In this context, brains are parts of agents that help process information in order to propagate those agents. From this abstract perspective, we can see that such ability is not restricted to brains \citep{Levin2020}. Thus, there are other mechanisms capable of producing intelligent behavior.

\section{Swarms}

There has been much work related to collective intelligence and cognition \citep{Hutchins1995,Heylighen1999,Reznikova:2007,Couzin:2009,malone2015handbook,SOLE201647}. Interestingly, groups of humans, animals or machines do not have a single brain. Thus information processing is distributed. 

A particular case is that of insect swarms \citep{Chialvo:1995,Garnier:2007,Passino:2008,Marshall:2009,Trianni-Tuci:09:ecal,Martin:2010}, where not only information processing is distributed, but also reproduction and selection occur at the colony level \citep{Holldobler2008}. 

To compare the cognitive architectures of brains and swarms, I previously proposed \emph{computing networks} \citep{Gershenson:2010b}. With this formalism, it can be shown that the differences in substrate do not necessarily imply a theoretical difference in cognitive abilities. Nevertheless, in practice, the speed and scalability of information processing of brains is much superior than that of swarms.

Thus, the brain as computer metaphor is not appropriate for studying collective intelligence in general, nor swarm intelligence in particular. However, the intelligence of brains and swarms can be  described in terms of information processing, as an agent $a$ can be an organism or a colony, with its own satisfaction $\sigma$ defined by an external observer.

Another advantage of studying intelligence as information processing is that we can use the same formalism to study intelligence at multiple scales: cellular, multicellular, collective/social, and cultural. Curiously, at the global scale, the brain metaphor has also been used \citep{mayer1995global,borner2005studying,Bernstein2012GlobalBrain}, although its usefulness remains to be demonstrated.

\section{Evolution and ecology}

If we want to have a better understanding of intelligence, we must study how it came to evolve. Intelligence as information-processing can also be useful in this context, as different substrates and mechanisms can be used to exhibit intelligent behavior. 

What could be the ecological pressures that promote the evolution of intelligence? Since environments and ecosystems can also be described in terms of information, we can say that more \emph{complex} environments will promote --- through natural selection --- more complex organisms and species, which will require a more complex intelligence to process the information of their environment and of other organisms and species they interact with \citep{Gershenson:2007}. In this way, the complexity of ecosystems can also be expected to increase though evolution. 

These ideas generalize Dunbar's \citeyearpar{Dunbar1993,Dunbar2003} ``social brain hypothesis'': larger and more complex social groups put a selective pressure on more complex information processing (measured as the neocortex to bodymass ratio), which gives individuals more cognitive capacities to recognize different individuals, remember who can they trust, multiple levels of intentionality \citep{dennett1989intentional}, and so on. In turn, increased cognitive abilities lead to more complex groups, so this cycle reinforces the selection for more intelligent individuals. 

One can make a similar argument using environments instead of social groups: more complex ecosystems put a selective pressure for more intelligent organisms, social groups, and species; as they require greater information-processing capabilities to survive and exploit their environments. This also creates a feedback, where more complex information processing by organisms, groups, and species produce more complex ecosystems. 

However, individuals can ``offload'' their information processing to their group or environment, leading to a decrease in their individual information processing abilities \citep{10.3389/fnbot.2021.634085}. This is to say that intelligence does not always increase. Although there is a selective pressure for intelligence, its cost imposes limits that depend as well on the usefulness of increased cognitive abilities.

Generalizing, we can say that information evolves to have greater control over its own production \citep{Gershenson:2007}. This leads to more complex information-processing, and thus, we can expect intelligence to increase at multiple scales through evolution, independently on the substrates that actually do the information processing. 

Another way of describing the same: information is transformed by different causes. This generates a \emph{variety} of complexity \citep{Ashby1956,Gershenson2015Requisite-Varie}. More complex information requires more complex agents to propagate it, leading to an increase of complexity and intelligence through evolution.

At different scales, since the Big Bang, we have seen an increase of information processing through evolution. In recent decades, this increase has been supraexponential in computers \citep{591665}. Although there are limitations for sustaining this rate of increase \citep{doi:10.1098/rsta.2019.0061}, we can say that the increase of intelligence is a natural tendency of evolution, be it of brains, swarms, or machines. This will not lead to a ``singularity'', but to an increase of the intelligence and complexity of humans, machines, and the ecosystems we create.

\section{Conclusion}

Brains are not essential for intelligence. Plants, swarms, bacterial colonies, robots, societies, and more exhibit intelligence without brains. An understanding of intelligence (and life \citep{Gershenson2020}) independently of its substrate, in terms of information processing, will be more illuminating that focussing only on the mechanisms used by vertebrates and other animals. In this sense, the metaphor of the brain as a computer, is limited more on the side of the brain than on the side of the computer. Brains do process information to exhibit intelligence, but there are several other mechanisms that also process information to exhibit intelligence. Brains are just a particular case, and we can learn a lot from them, but we will learn more if we do not limit our studies to their particular type of cognition.

\section*{Conflict of Interest Statement}

The authors declare that the research was conducted in the absence of any commercial or financial relationships that could be construed as a potential conflict of interest.

\section*{Author Contributions}

C.G. conceived and wrote the paper.


\section*{Funding}
This work was supported by UNAM's PAPIIT IN107919 and IV100120 grants.




\bibliographystyle{frontiersinSCNS_ENG_HUMS} 
\bibliography{carlos,refs}

\end{document}